\documentclass[11pt]{article}

\usepackage[preprint]{acl}

\usepackage{times}
\usepackage{latexsym}

\usepackage[T1]{fontenc}
\usepackage[utf8]{inputenc}
\usepackage{microtype}
\usepackage{inconsolata}
\usepackage{graphicx,subcaption}
\usepackage{amsmath}
\usepackage{amssymb}
\usepackage{amsthm}
\usepackage{booktabs}
\usepackage{longtable}
\usepackage{nameref}
\usepackage{tabularx}
\usepackage{caption} 
\usepackage{multicol}
\usepackage{paracol}

\usepackage{float}

\newtheorem{proposition}{Proposition}

\title{Hidden Consensus: \\
Preference-Validity Compression in Human Feedback}

\author{Dorcas Chia Ern Chua\textsuperscript{1} \qquad Karen Myn Hui Lee\textsuperscript{1} \qquad Jia Yue Tan\textsuperscript{1} \qquad Zhen Xue Gue\textsuperscript{3,$\circ$}\thanks{Intern at Universiti Malaya.} \\
\textbf{ \qquad Norzalena Abdul Hamid\textsuperscript{1} \qquad Azima Binti Azmi\textsuperscript{1} \qquad Keat Mei Yeong\textsuperscript{1}} \\
\textbf{Aizat Izyani binti Mujab\textsuperscript{1} \qquad Hafsah Noor Azam\textsuperscript{1} \qquad Chee Guo Khoo\textsuperscript{4,$\circ$}\thanks{Intern at YTL AI Labs.}} \\
\textbf{Han Ying Lim\textsuperscript{1,$\diamondsuit$} \qquad Chee Seng Chan\textsuperscript{2,$\diamondsuit$}} \\
\\
  \textsuperscript{$\circ$}Intern \qquad
  \textsuperscript{$\diamondsuit$}Project Lead
\\
  \textsuperscript{1}YTL AI Labs\, 
  \textsuperscript{2}Universiti Malaya\, 
  \textsuperscript{3}Monash University Malaysia\,
  \textsuperscript{4}Universiti Malaysia Sarawak 
}

\begin{document}
\maketitle

\begin{abstract}
Standard RLHF pipelines often reduce heterogeneous human judgments into a single scalar reward target. We argue that this reduction can mis-measure alignment in structurally plural societies, where disagreement may reflect culturally, historically, linguistically, regionally, or normatively grounded interpretations rather than annotation noise. We call this failure \emph{Preference-Validity Compression}, the collapse of multiple  plural-valid response options into a single optimization target. Using Malaysia as a diagnostic setting, we analyze RLHF-style feedback aggregation through \emph{preference events} linking prompts, responses, and acceptability judgments across interpretive frames. Across 321 preference events from 20 participants and 107 trio-annotated prompts, 79\% of prompts contain more than one majority-supported response that single-winner aggregation would discard, and apparent dominance gaps between top responses diminish when all majority-supported options are considered. Participants frequently select multiple acceptable responses, and discarded responses demonstrably reflect coherent local, practical, or cultural frames. These findings show that majority aggregation in this corpus measures $\arg\max$ acceptability rather than plural alignment. We treat this as a measurement-validity issue and argue that future alignment methods should satisfy \emph{Validity-Preserving Consistency}, remaining stable across plural-valid interpretive frames rather than collapsing them into a single reward target.
\end{abstract}

%%%%%%%%%%%%%%%%%%%%%%%%%%%%%%%%%%%%%%%%%%%%%%%
\section{Introduction}

Reinforcement learning from human feedback (RLHF) aligns large language models by converting human judgments over model outputs into a scalar reward signal \citep{christiano2017deep,ouyang2022training}. This scalarization is useful for optimization, but it also creates a measurement problem. Standard reward modeling often treats pooled preference comparisons as observations of a shared latent preference function, with disagreement absorbed as stochastic variation. This assumption is reasonable when disagreement reflects noise or annotation error. It becomes fragile when disagreement reflects more than one valid interpretation of what constitutes an acceptable response.

\begin{figure}[t]
  \includegraphics[width=\linewidth, keepaspectratio]{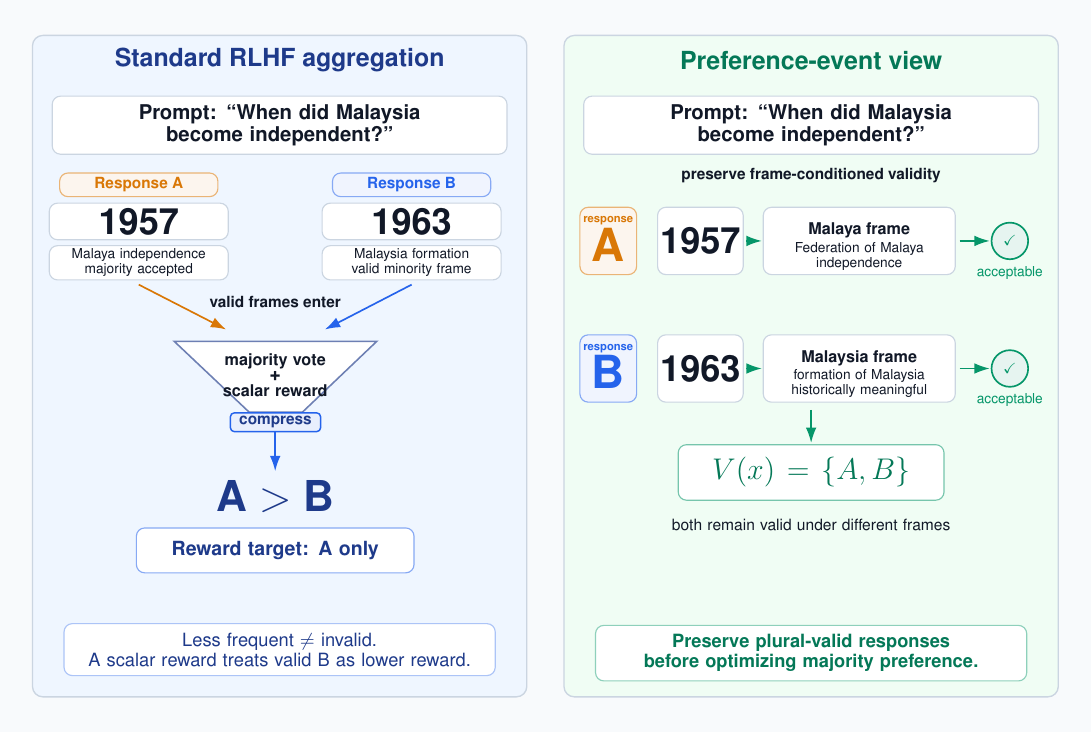}
  \caption{Preference-Validity Compression. Single-winner aggregation selects ``1957'' as the reward target, while the preference-event view preserves both ``1957'' and ``1963'' as acceptable under different frames.}
  \label{fig:intro-boundary}
  \vspace{-1em}
\end{figure}

This distinction matters for pluralistic alignment. Prior work shows that annotator disagreement can encode meaningful semantic, moral, and cultural variation rather than error \citep{aroyo2015truth,sorensen2024position}. Other work shows that aggregated reward models can converge toward dominant preference patterns and encode narrow annotator populations as if they were universal preferences \citep{casper2023open,santurkar2023whose}. We study this as an alignment-measurement problem. The source of the failure is not disagreement itself, but the aggregation step that turns heterogeneous acceptability judgments into one reward target. When feedback is reduced to a single target, the system does not only summarize human preference. It determines which acceptable responses remain visible to optimization.

We call this failure \emph{Preference-Validity Compression}. It occurs when multiple acceptable response options are collapsed into one dominant optimization target. Figure~\ref{fig:intro-boundary} illustrates the idea using the question of Malaysia's independence. The 1957 frame reflects the independence of Federation of Malaya, while 1963 frame reflects the formation of Malaysia, involving Malaya, Sabah, Sarawak, and Singapore at the time. A single reward target may preserve one frame while suppressing another frame that remains historically meaningful.

We use Malaysia as a diagnostic setting because it is structurally plural. Disagreement over acceptable model responses may reflect coexisting cultural, historical, linguistic, religious, and regional interpretive frames rather than random annotator noise. The category “Malaysian users” should not be assumed to correspond to a single preference signal. Our goal is not to estimate national preference distributions or train a reward model, but to test whether single-winner aggregation can conceal supported alternatives within a structurally plural setting.

We analyze human feedback as \emph{preference events}, context-dependent judgments formed by a prompt, a model response, and an evaluator's acceptability decision mediated by intersecting cultural, historical, linguistic, and normative frames. In our diagnostic study, 20 Malaysian participants judged three responses per prompt as acceptable or not acceptable across 107 trio-annotated prompts. This elicitation design allows plural acceptability to surface prior to aggregation and enables us to test whether the supported response set is non-singleton at the event level.

The results show that this measurement issue is observable in practice. Across 
321 preference events, 79\% of prompts contain at least one additional response 
that reaches the majority acceptance threshold but would be discarded by 
$\arg\max$ aggregation. Moreover, single-winner aggregation overstates the 
dominance of the winning response: gaps between top responses diminish substantially when all majority-supported responses are considered. At the annotator level, acceptance behavior also does not conform to a stable one-response-per-prompt pattern, suggesting that evaluators frequently treat multiple responses as simultaneously valid rather than mutually exclusive. Qualitative hidden-consensus examples further show that discarded responses can reflect coherent local, practical, or cultural frames.

These findings reframe pluralistic alignment as a measurement-validity problem where 
the aggregation protocol itself becomes the hidden bottleneck when it collapses 
plural-valid interpretations into a single reward target. We argue that future 
alignment methods should satisfy \emph{Validity-Preserving Consistency}. Alignment should 
remain stable across plural-valid interpretive frames rather than collapsing 
them into a dominant reward target.

This paper makes three contributions.
\begin{itemize}
    \item We identify a measurement failure in RLHF-style feedback aggregation, where a single reward target can measure $\arg\max$ acceptability while being interpreted as plural alignment.
    \item We formalize this failure as \emph{Preference-Validity Compression} 
    and introduce \emph{preference events} as the diagnostic unit linking 
    acceptability judgments to prompts, responses, interpretive frames, and 
    evaluator context.
    \item Using Malaysia as a diagnostic setting, we provide empirical evidence that plural acceptability persists at both the response-set and annotator levels, causing single-winner aggregation to mis-measure the breadth and structure of supported responses in this corpus.
\end{itemize}
%%%%%%%%%%%%%%%%%%%%%%%%%%%%%%%%%%%%%%%%%%%%%%%
\section{Related Work}

\subsection{Annotator Disagreement and Preference Aggregation}

NLP research has challenged the view that annotator disagreement is merely measurement error. Disagreement can reflect semantic ambiguity, subjective interpretation, and value pluralism rather than noise \citep{aroyo2015truth,basile2021consider}. For RLHF, aggregating plural-valid disagreement determines which interpretations remain visible to the learning system.

Standard RLHF pipelines typically collect human preference comparisons, fit a reward model to pooled judgments, and optimize against the learned scalar reward. Recent work challenges this single-reward assumption. PRISM shows that alignment preferences are subjective and context-dependent across 1,500 participants from 75 countries \citep{kirk2024prism}. Heterogeneous RLHF and MaxMin-RLHF study personalization, aggregation, and group-aware objectives for diverse preferences \citep{park2024rlhf,chakraborty2024maxmin}. Operationalizing Pluralistic Values further shows that demographic composition, disagreement handling, rating scales, and optimization choices can change learned model behavior \citep{ali2025operationalizing}. Prior work on annotation aggregation 
shows that majority voting can under-represent sociodemographic groups and embed 
representational bias into downstream models \citep{prabhakaran2021releasing}, 
and social-choice perspectives treat preference aggregation as a normative 
design choice rather than a neutral step \citep{conitzer2024social}. These 
works establish that preference heterogeneity matters for alignment, but 
diversity-aware alignment can still become non-plural if it ultimately collapses 
acceptable alternatives into one selected target. What remains unaddressed is 
what the aggregation target actually measures when multiple responses are 
acceptable for the same prompt.

\subsection{Cross-Cultural Alignment and Malaysian Representation}

Concerns about narrow preference representation build on critiques of WEIRD bias, where narrow populations are often treated as default human subjects \citep{henrich2010weirdest}. Recent alignment work argues for broader community participation \citep{mihalcea2025weird}. For RLHF, however, inclusion alone is insufficient. A diverse annotator pool can still produce a non-plural reward model if disagreement is collapsed into one scalar target.

Empirical work shows that LLM behavior often reflects dominant cultural frames. Multilingual LLMs may produce responses closer to English-speaking cultural contexts even when queried in other languages, and this tendency can increase after RLHF-style alignment \citep{wang2024thanksgiving}. From our perspective, this pattern arises only partly from issues in pretraining coverage. When the reward target privileges one dominant frame, other locally valid interpretations are rendered invisible. The deeper alignment risk thus extends beyond issues of under-representation, as supported non-winner interpretations may disappear entirely when aggregation collapses plural acceptability into a single optimization target. 

Recent Southeast Asian benchmarks show that Malaysian linguistic and cultural representation remains challenging. MalayMMLU studies Malaysian knowledge evaluation in Bahasa Melayu \citep{poh2024malaymmlu}, while MyCulture examines Malaysian cultural knowledge and the risk of treating nations as homogeneous cultural units \citep{hew2025myculture}. These benchmarks ask whether models know Malaysian facts; we ask whether human feedback can preserve plural 
Malaysian acceptability judgments. A model may know the relevant facts while 
still being aligned to only one dominant evaluative frame --- a failure the 
benchmarking literature does not currently measure.

%%%%%%%%%%%%%%%%%%%%%%%%%%%%%%%%%%%%%%%%%%%%%%%
\section{Formalizing Preference-Validity Compression}

Let $x$ denote a prompt and let $Y_x=\{y_1,\ldots,y_m\}$ denote the candidate responses. Let $u_i(x,y)\in\{0,1\}$ indicate whether evaluator $i$ judges response $y$ acceptable for $x$. Because evaluators may accept multiple responses, $u_i(x,y)$ is not a forced-choice ranking. Standard scalar aggregation estimates acceptability as
\begin{equation}
A(x,y)=\frac{1}{n}\sum_{i=1}^{n}u_i(x,y),
\label{eq:acceptability}
\end{equation}
and a single-winner objective selects
\begin{equation}
y^* \in \arg\max_{y\in Y_x} A(x,y).
\label{eq:majority-objective}
\end{equation}
When multiple responses tie, $y^*$ denotes one selected winner under an arbitrary or implementation-specific tie-breaking rule.

Pluralistic alignment requires a different measurement object. Let $V(x)\subseteq Y_x$ denote the set of plural-valid responses for $x$. A response belongs to $V(x)$ if it is acceptable under a legitimate cultural, historical, linguistic, regional, or normative frame, even when it is not the top-ranked response. Thus, $V(x)$ is a validity set, not a frequency ranking. In our empirical study, we do not observe $V(x)$ directly. We diagnose it through majority-supported acceptability patterns and qualitative justifications.

We define \emph{Preference-Validity Compression} as the failure mode in which scalar aggregation replaces the plural-valid set $V(x)$ with the single-winner target $y^*$. The failure is not that a lower-frequency response is always valid. The failure is that scalar aggregation alone cannot distinguish an invalid non-winner from a valid non-winner. A response may disappear from the optimization target not because it is invalid, but because the aggregation operator retains only one selected response.

\begin{proposition}
If a prompt admits more than one plural-valid response, a single-winner majority objective cannot preserve plural validity as a set. It maps $V(x)$ to one response and omits valid non-winner responses from the optimization target.
\end{proposition}

The proof is given in Appendix~\ref{app:proof}. The proposition identifies a measurement limitation rather than a claim that every non-winner response is valid. A single scalar ranking does not reveal whether a non-winner response is invalid or valid under another frame. Our empirical study tests this failure through three signals in Section~\ref{sec:results}, that are accepted-response multiplicity, majority-compression loss, and non-fixed participant modes.

%%%%%%%%%%%%%%%%%%%%%%%%%%%%%%%%%%%%%%%%%%%%%%%
\section{Malaysia as a Diagnostic Context}

We use Malaysia as a diagnostic context for Preference-Validity Compression because its structural plurality challenges the assumption that pooled population feedback can be reduced to a single preference signal. Malaysian society is commonly described through broad ethnolinguistic categories: "Malay", "Chinese", "Indian", and "Other", which encompasses diverse indigenous communities across Peninsular Malaysia, Sabah, and Sarawak \citep{dosm2026demographic}. These categories are historically constructed, but remain socially and institutionally salient in education, economic organisation, language policy, and everyday interaction \citep{hirschman1986making,shamsul1996debating,shamsul2009stable,ting2014race}. 

We do not argue that each group has a fixed preference profile, nor do we aim to estimate population-level preference distributions. Rather, Malaysia is a useful diagnostic context because disagreement reflects coexisting, legitimate interpretive frames, not random annotator noise. This aligns with descriptions of Malaysia as a ``state in stable tension'', where social order is maintained through negotiation among competing social and political claims instead of settled consensus \citep{shamsul2009stable}. Thus, feedback aggregation using a single reward target does not recover any hidden national preference but instead determines which interpretive frame remains visible to optimization.

The Merdeka Center's 2024 National Youth Survey illustrates this structure.The national sample is nearly evenly divided on the issue of ethnic-based Bumiputera privileges versus equal treatment across race and religion, with 49\% supporting the former and 48\% favouring the latter \citep{merdeka2024youth}. Beneath this near-parity, however, are divergent regional and ethnic patterns: 73\% of Malay youth respondents support continued Bumiputera privileges while 65\% of East Malaysian respondents support equal treatment \citep{merdeka2024youth}. The point is not that ethnicity or region deterministically predicts preference. More precisely, value-laden judgments, which are shaped by different cultural, historical, regional
and normative frames, can remain socially intelligible within the same national context.

%%%%%%%%%%%%%%%%%%%%%%%%%%%%%%%%%%%%%%%%%%%%%%%
\section{Study Design}

We present a controlled diagnostic study of model response acceptability in Malaysia. The study tests whether locally relevant prompts can admit multiple acceptable responses, and whether single-winner aggregation would suppress majority-supported non-winner responses by diagnosing the feedback-elicitation and aggregation step that precedes reward optimization.

\subsection{Participants and Prompt Assignment}

We recruited participants through an online registration form and selected them to provide coverage across Malay, Chinese, Indian, and indigenous groups from Sabah and Sarawak, across age bands from 18-24 to 65 and above. Participants were compensated in accordance with Malaysian minimum wage requirements. Participants provided informed consent for their responses to be used in this research; Full details on participant recruitment can be found in Appendix~\ref{app:recruitment}. 

Each prompt was assigned to three participants, allowing a minimal majority judgment of two out of three. Assignment was random rather than demographically stratified within prompt groups.

\begin{figure*}[t]
  \centering
  \includegraphics[width=\linewidth]{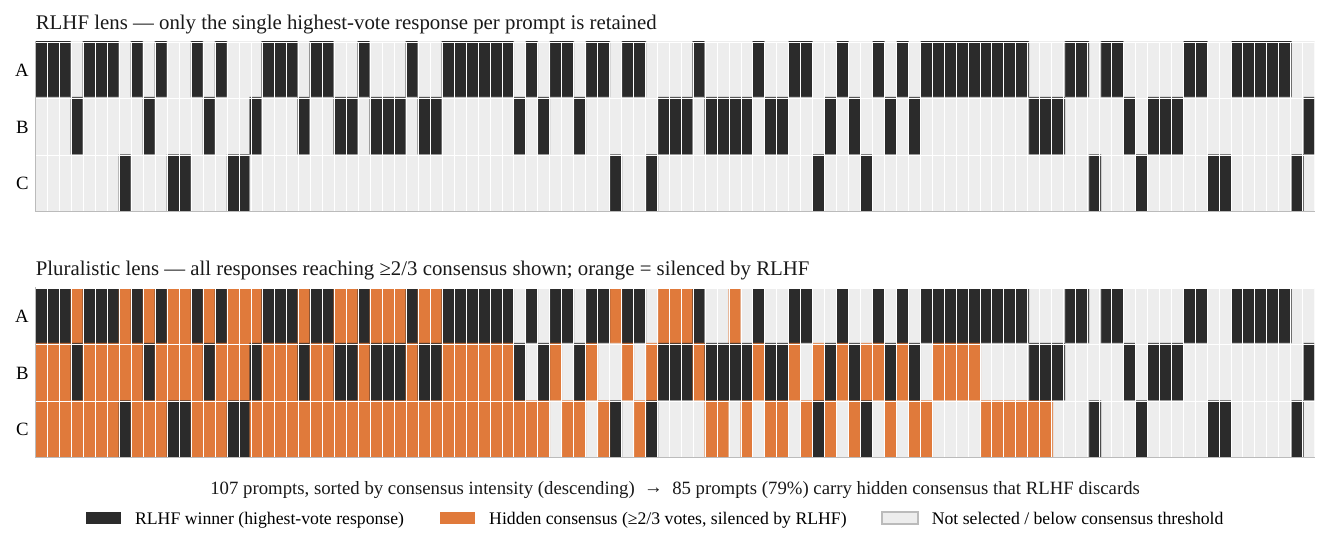}
  \caption{What majority aggregation retains compared with the full set of $\geq 2/3$ majority-supported responses across 107 trio-annotated prompts. Each column is one prompt. Dark cells indicate the $\arg\max$ response a single-winner reward model would optimize toward. Orange cells indicate additional responses reaching $\geq 2/3$ acceptance that majority aggregation discards as hidden consensus. Overall, 85 of 107 prompts, or 79\%, contain at least one such discarded response.}
  \label{fig:rlhf-vs-pluralistic}
  \vspace{-1em}
\end{figure*}

\subsection{Prompt and Response Construction}

Prompts were grounded in Malaysian online discourse to avoid relying only on researcher assumptions about local disagreement. We systematically collected and processed content from four Malaysian digital platforms, such as Lowyat.net, Reddit, Threads, and YouTube podcast channels producing 692,919 utterances across 26,463 source files after segmentation, filtration, and deduplication. Recurring topics from this corpus were used to seed prompt generation with DeepSeek-V4-Flash. The prompts cover value-laden situations where cultural, religious, linguistic, or regional background may plausibly affect what counts as an acceptable response. The initial corpus contained 371 prompts before quality filtering. Full details are provided in Appendix~\ref{sec:prompt-generation}.

\begin{figure*}[t]
  \centering
  \begin{subfigure}[t]{0.48\textwidth}
    \centering
    \includegraphics[width=\linewidth]{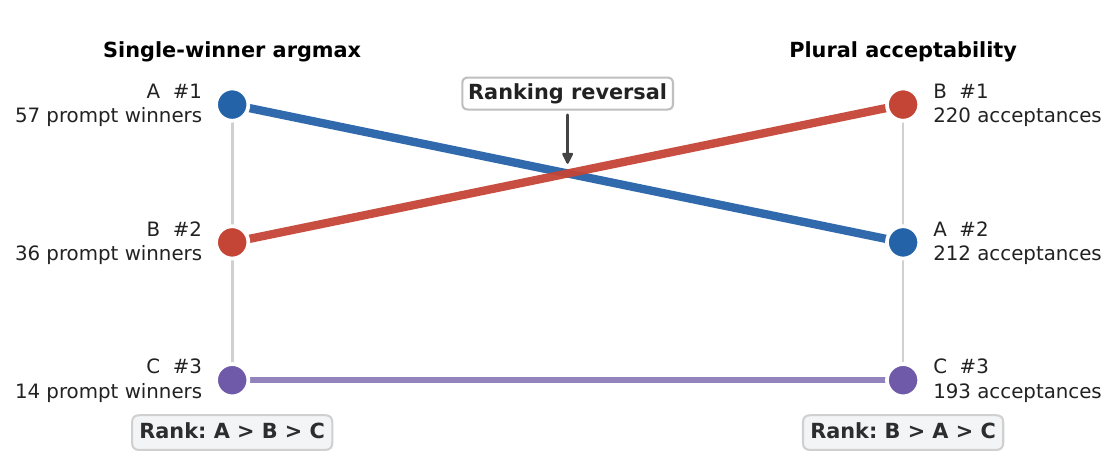}
    \caption{Ranking reversal}
    \label{fig:argmax-compression-ranking}
  \end{subfigure}
  \hfill
  \begin{subfigure}[t]{0.48\textwidth}
    \centering
    \includegraphics[width=\linewidth]{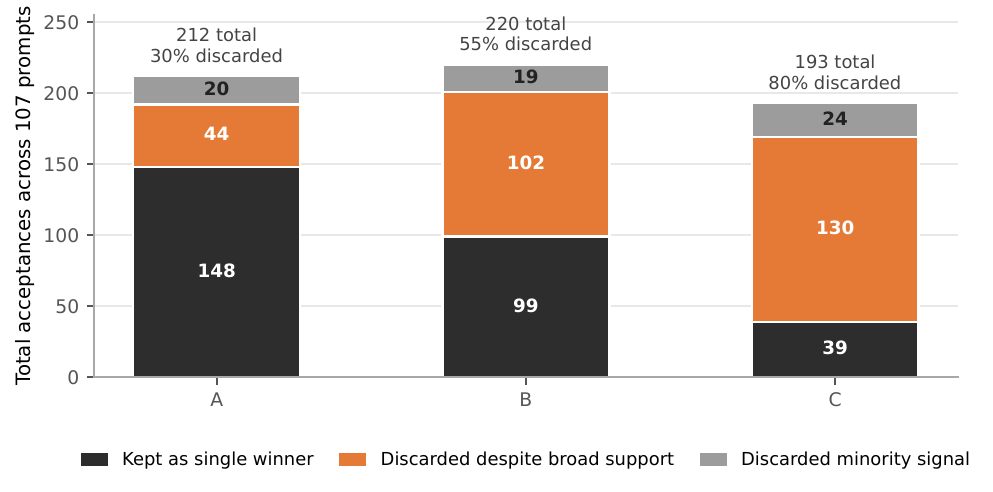}
    \caption{Discarded support}
    \label{fig:argmax-compression-discarded}
  \end{subfigure}
  \caption{$\arg\max$ compression hides plural acceptability. (a) Under single-winner $\arg\max$, response $A$ appears dominant with 57 prompt-level wins, followed by $B$ with 36 and $C$ with 14. Under the majority-threshold view, $A$ and $B$ are mostly tied at 79-80 prompts, while $C$ remains broadly supported in 73 prompts. (b) At the acceptance-count level, total support ranks responses as $B>A>C$, and many accepted responses are discarded by single-winner aggregation.}
  \label{fig:argmax-compression}
  \vspace{-5pt}
\end{figure*}

For each prompt, three candidate responses were generated using DeepSeek-V4-Flash, GPT-5-Nano, and Gemini-3.1-Flash-Lite-Preview. Responses were intended to represent agreement, disagreement, and balanced positions, but this intended stance scheme was not always realized. We therefore treat response positions $A$, $B$, and $C$ as empirical candidates rather than controlled stance conditions. Model identities were not disclosed to participants. More details are included in Appendix~\ref{sec:response-generation}.

\subsection{Acceptability Elicitation and Filtering}

Instead of forced-choice preference selection, participants rated each of the three responses independently as acceptable or not acceptable and provided a brief written justification for each rating. This format allows participants to accept exactly one response, accept more than one response, or reject all responses. By removing the forced-choice constraint, the design makes Preference-Validity Compression diagnosable. If plural acceptability cannot appear in the data, it cannot be measured.

After data collection, we applied two filters. First, prompts were removed when the scenario was internally inconsistent or factually incoherent. Second, prompts without complete responses from all three assigned participants were removed, ensuring that every retained prompt supported a two-out-of-three majority judgment. The final analysable dataset comprises 20 participants, 107 prompts, 321 preference events, and 963 response-level ratings.

%%%%%%%%%%%%%%%%%%%%%%%%%%%%%%%%%%%%%%%%%%%%%%%
\section{Results and Analysis}
\label{sec:results}

We analyze $N=321$ preference events from 20 participants across 107 trio-annotated prompts. Each event records one participant's acceptability decision over three candidate responses, expressed as a binary selection vector over $\{A,B,C\}$. Across 963 possible response-level ratings, participants marked 625 responses as acceptable. The analysis tests three diagnostics introduced in Section~3, which are the accepted-response multiplicity, majority-compression loss, and non-fixed participant modes. The goal is not to simulate a full RLHF pipeline, but to examine whether majority-supported acceptability distributes across multiple responses in ways that a single-winner reward target would compress.

\subsection{Accepted-Response Multiplicity}

The first diagnostic asks whether a prompt admits more than one acceptable response. Under a single-preference latent utility assumption, acceptability should concentrate around one $y^\ast$ per prompt, with other acceptances treated as residual variation. A plural-acceptability view predicts that more than one response may receive substantial support when responses instantiate coexisting interpretive frames.

For each prompt, we count the number of responses accepted by at least two of three participants, the minimum majority threshold in this annotation design. Figure~\ref{fig:rlhf-vs-pluralistic} contrasts the $\arg\max$ response retained by single-winner aggregation with the full set of responses reaching $\geq 2/3$ acceptance. We call additional majority-supported non-winners \emph{hidden consensus}.

Across 107 prompts, 85 prompts, or 79.4\%, admit two or more responses reaching $\geq 2/3$ acceptance. Only 21 prompts, or 19.6\%, exhibit a single majority-supported response, and one prompt, or 0.9\%, produces no majority-supported response. Thus, the corpus is not characterized by sparse or noisy consensus. It is characterized by coexisting consensus. The reduction $y^\ast=\arg\max_y A(x,y)$ retains only one response and discards other responses that meet the same majority threshold. In most prompts, the empirically supported acceptability set is non-singleton and cannot be recovered from $y^\ast$ alone.

\subsection{Majority-Compression Loss}

The second diagnostic asks what is lost when acceptability is reduced to a single $\arg\max$ target. Figure~\ref{fig:argmax-compression} shows that the loss is not only quantitative but interpretive, where a single-winner aggregation changes which response appears most supported.

At the prompt level, Figure~\ref{fig:argmax-compression-ranking} shows that single-winner aggregation changes the apparent support structure. Under $\arg\max$ aggregation, response $A$ appears dominant, serving as the single-winner response in 57 prompts, followed by $B$ in 36 prompts and $C$ in 14 prompts. However, when every response reaching the $\geq 2/3$ majority acceptance threshold is counted, $A$ and $B$ each reach majority support in 79 prompts and 80 prompts respectively, while $C$ reaches majority support in 73 prompts. Thus, the $\arg\max$ view makes $A$ appear clearly dominant, whereas the majority-threshold view reveals a much flatter support structure. $A$ and $B$ are mostly tied, and $C$ is far less marginal than it appears under single-winner aggregation.

\begin{table*}[t]
\centering
\caption{Examples of hidden consensus discarded by single-winner aggregation. Each hidden response reached the majority threshold but would be removed by $\arg\max$ aggregation.}
\scriptsize
\begin{tabular}{p{0.18\linewidth}p{0.18\linewidth}p{0.18\linewidth}p{0.32\linewidth}}
\toprule
Prompt context & Kept by $\arg\max$ & Hidden consensus discarded & Why the hidden response was accepted \\
\midrule
Gaza workplace solidarity & B, pragmatic advice, 3/3 & C, pluralist balancing frame, 2/3 & Balances moral principle with career obligations and gives realistic options. \\
Bahasa Melayu in private company & B, pragmatic corporate frame, 3/3 & C, pluralist language-identity frame, 2/3 & Explains the tension between national language identity and private-sector communication norms. \\
Kampung proverb and local culture & A, pragmatic cultural advice, 3/3 & B, local-cultural interpretation, 2/3 & Interprets the proverb as a gentle cultural invitation rather than a personal insult. \\
Class bias in ``gaji buta'' discourse & A, institutional bias critique, 3/3 & B and C, pragmatic and pluralist critiques, 2/3 each & Addresses the class bias and double standard in how ``gaji buta'' is applied. \\
\bottomrule
\end{tabular}
\label{tab:hidden-consensus-examples}
\vspace{-10pt}
\end{table*}

\begin{table}[t]
\centering
\scriptsize
\begin{tabular}{p{0.48\linewidth}p{0.40\linewidth}}
\toprule
Robustness check & Finding \\
\midrule
No unique $\arg\max$ & 44/107 prompts, 41.1\% \\
Strict $3/3$ threshold & 23/107 prompts, 21.5\%, have multiple unanimously accepted responses \\
Language coverage & Hidden consensus appears in both BM and English prompts \\
Domain coverage & Appears across governance, language identity, religion, labour, and privacy \\
Response-position coverage & $A$, $B$, and $C$ reach $\geq 2/3$ in 79, 80, and 73 prompts \\
Participant behavior & All 20 participants average $>1$ accepted response per prompt, 19 use at least two selection patterns \\
\bottomrule
\end{tabular}
\caption{Robustness checks showing that hidden consensus is not driven by a single threshold, language subset, topic domain, response position, or participant habit.}
\label{tab:robustness}
\vspace{-10pt}
\end{table}

At the acceptance-count level, Figure~\ref{fig:argmax-compression-discarded} decomposes total acceptances into support retained by $\arg\max$ and support discarded by single-winner aggregation. Response $B$ receives the highest total support with 220 acceptances, followed by $A$ with 212 and $C$ with 193. Yet only 99, 148, and 39 of these acceptances respectively map onto the single-winner signal for $B$, $A$, and $C$. Equivalently, the discarded fraction is 55\% for $B$, 30\% for $A$, and 80\% for $C$. Response $C$, which appears marginal under $\arg\max$, loses four out of every five acceptances to compression.

This is the measurement-validity consequence formalized in Proposition~1. The $\arg\max$ reward signal does not merely under-weight weak minority responses. It discards majority-supported alternatives and can invert which responses appear most broadly endorsed. The failure is therefore not only loss of diversity, but loss of measurement validity, where majority aggregation measures $\arg\max$ acceptability, not plural alignment.

To check that hidden consensus is not merely indiscriminate acceptance, we inspect examples where a non-winner response reached the $\geq 2/3$ majority threshold but was discarded by $\arg\max$ aggregation. Table~\ref{tab:hidden-consensus-examples} shows that these discarded responses can correspond to coherent local, practical, or cultural frames. The examples are illustrative rather than exhaustive, but they show what single-winner aggregation removes at the level of interpretable response frames.

The robustness checks in Table~\ref{tab:robustness} support the interpretation that hidden consensus is a broad measurement pattern rather than a single-threshold or single-response artifact.

\subsection{Non-Fixed Participant Modes}

The third diagnostic asks if acceptability behaves as a fixed user-level trait or as a context-dependent preference event. Under a fixed single-selection view, participants should usually accept only one response per prompt. Instead, multi-response selection is the dominant empirical pattern.

\begin{figure}[t]
  \centering
  \includegraphics[width=\linewidth]{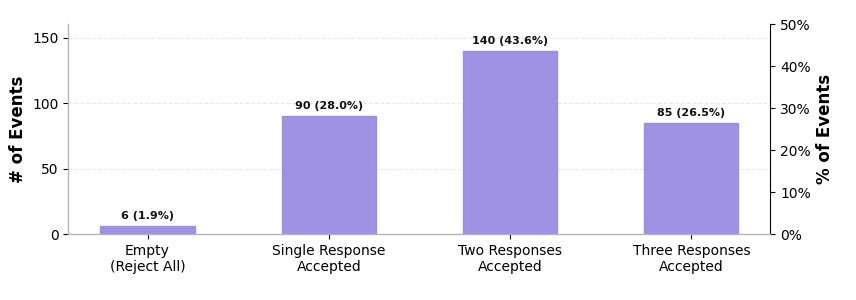}
  \caption{Selection-pattern distribution across 321 preference events. Multi-response selection is the modal behavior. Two-response selections account for 140 events, or 43.6\%, while accepting all three responses accounts for 85 events, or 26.5\%. Single-response selection accounts for 90 events, or 28.0\%, and rejecting all responses accounts for 6 events, or 1.9\%.}
  \label{fig:selection-pattern-distribution}
  \vspace{-15pt}
\end{figure}

Figure~\ref{fig:selection-pattern-distribution} shows that plural acceptability is visible at the participant-event level, not only after prompt-level aggregation. Across 321 events, 225 events, or 70.1\%, involve accepting more than one response. Two-response patterns account for 140 events, accepting all three responses accounts for 85 events, and single-response patterns account for only 90 events. This supports the preference-event view, where participants often judged more than one response acceptable for the same prompt.

A natural objection is that multi-selection reflects indiscriminate acceptance. We test this in Appendix~\ref{app:participant-mode}. Participant-level selection breadth ranges from 1.33 to 2.69 responses per prompt, while consensus alignment rates remain high, ranging from 71\% to 100\%. We find no detectable linear association between selection breadth and consensus alignment rate, with Pearson $r=-0.09$ and $p=0.71$. Participants who accept more responses are therefore not systematically landing on less-supported responses. They are surfacing additional majority-supported responses that single-winner aggregation leaves behind.

Taken together, the three diagnostics show that Malaysian human feedback in this corpus cannot be faithfully represented as a single per-prompt majority preference. Majority-supported plural acceptability arises in 79\% of prompts. $\arg\max$ compression silences four out of every five acceptances for the response that appears marginal under majority aggregation. Multi-selection is the modal annotation behavior, with no detectable consensus-support penalty for participants who accept multiple responses. Majority aggregation therefore measures the most frequently selected response, not plural alignment. RLHF-style feedback aggregation in structurally plural settings should preserve $V(x)$ as a set rather than collapse it into $y^\ast$.

%%%%%%%%%%%%%%%%%%%%%%%%%%%%%%%%%%%%%%%%%%%%%%%
\section{Discussion}
Our results show that Preference-Validity Compression is observable in a diagnostic human evaluation corpus. The central implication is not simply that annotators disagree. It is that RLHF-style feedback aggregation can change what construct is being measured. For instance, from plural acceptability to single-winner acceptability.

\paragraph{Plural acceptability is the modal empirical pattern.}
The 79\% multiplicity rate shows that, for most prompts in this corpus, a single-winner reward objective would discard at least one response that met the same majority threshold as the retained winner. Multi-selection was also the modal annotation behavior across 321 preference events. Participants who accepted more responses were not accepting less consensus-supported ones; they were surfacing additional responses that also commanded majority support. Plural acceptability is therefore not a marginal deviation around a dominant signal in this corpus. It is the dominant empirical pattern.

\paragraph{Majority aggregation changes the observed support structure.}
Compression loss (Section~\ref{sec:results}) shows that majority aggregation does not merely discard weak minority signals. It can also discard majority-supported non-winner responses and change which responses appear most broadly supported. Response $C$, which appears marginal under $\arg\max$ aggregation, loses four out of every five acceptances to compression. A single-winner reward target based on this signal will hence misrepresent support structure, not simply simplify it.

\paragraph{Preference-Validity Compression is not solved by sampling alone.}
A larger or more demographically stratified sample may refine the estimated distribution, but it does not by itself solve the measurement problem. The same participants produce different selection patterns across prompts, and selection breadth shows no detectable linear association with consensus alignment rate. The same applies to demographic diversification. A diverse annotator pool can still produce a non-plural reward model if disagreement is averaged away rather than preserved \citep{casper2023open,prabhakaran2021releasing}. The problem is therefore not only who is in the pool, but what the aggregation operator does to their judgments.

\paragraph{Forced-choice formats can manufacture the convergence RLHF assumes.}
Our free-acceptance format makes plural acceptability detectable. Standard forced-choice formats may not merely measure preference convergence; they may also produce it. When participants must select one response, the data will always contain a winner, and the elicitation instrument that produced this convergence disappears inside the reward signal. This does not invalidate RLHF as an optimization procedure, but it raises a prior measurement question about what its reward signal represents.

\paragraph{Implications for pluralistic alignment.}
Our contribution is diagnostic. We identify Preference-Validity Compression as a measurement-validity failure rather than propose a replacement algorithm. Existing approaches such as MaxMin-RLHF and personalized RLHF address preference heterogeneity by changing the optimization target or modeling user variation \citep{chakraborty2024maxmin,park2024rlhf,chen2024pal}. These directions are important, but they do not by themselves guarantee validity preservation. A method can model diverse preferences while still selecting a single response per prompt. Our results suggest that the bottleneck is not only who is in the annotator 
pool but what the aggregation operator does to their judgments. Pluralistic 
alignment should therefore preserve $V(x)$ as the set of responses valid 
under different interpretive frames, rather than collapse it into a single 
target $y^\ast$, a property we term \emph{Validity-Preserving Consistency}.

\paragraph{Fairness implication.}
From a fairness perspective, the harm is not only that diversity is reduced. When hidden consensus is discarded, valid minority or alternative frames are removed from the learning signal and become invisible to optimization. The resulting reward target may then reinforce dominant-frame responses while treating less frequent but still valid frames as lower-quality, less preferred, or less aligned.

%%%%%%%%%%%%%%%%%%%%%%%%%%%%%%%%%%%%%%%%%%%%%%%
\section{Conclusion}
This paper reframes RLHF-style feedback aggregation as a measurement-validity problem. In structurally plural settings, disagreement may reflect plural-valid interpretations rather than annotation noise, yet scalar aggregation can collapse these interpretations into one optimization target. We call this failure \emph{Preference-Validity Compression}. Using Malaysia as a diagnostic setting, we show that plural acceptability can appear before aggregation. Across 321 preference events from 20 participants and 107 trio-annotated prompts, 79\% of prompts contained more than one majority-supported response, multi-selection was the most frequent annotation behavior, and $\arg\max$ aggregation inverted the observed support ranking. These findings suggest that majority aggregation in this corpus measures $\arg\max$ acceptability rather than plural alignment. Future alignment methods should therefore satisfy \emph{Validity-Preserving Consistency}, preserving $V(x)$ as a set of plural-valid responses before optimizing a dominant preference.

%%%%%%%%%%%%%%%%%%%%%%%%%%%%%%%%%%%%%%%%%%%%%%%
\clearpage
\newpage

\section*{Limitations}
\label{sec:Limitations}

\paragraph{Diagnostic scope.}
This study is a controlled diagnostic rather than a nationally representative survey. The 20-participant final sample was selected to provide ethnolinguistic and age coverage, not to estimate population-level preference distributions in Malaysia. The findings should therefore be read as evidence that Preference-Validity Compression can arise under controlled plural-feedback conditions, not as a national prevalence estimate.

\paragraph{Sample size and prompt assignment.}
Each retained prompt was evaluated by three participants, which enables a minimal majority signal but limits the resolution of within-prompt disagreement. Because assignment was random and not demographically stratified, prompts on regionally or ethnically specific issues may have been judged by participants outside the relevant frame, which would understate hidden consensus. The reported 79\% is therefore likely a lower bound.

\paragraph{Stance labelling.}
Stance labels were assigned independently by five members of the research team without adjudicating disagreements into a single resolved label. The schema was developed iteratively during analysis. For this reason, stance labels are treated as descriptive characterisations of response content, not as ground-truth categorical measurements. The main quantitative results rely on acceptability counts rather than stance-label agreement.

\paragraph{Response generation.}
The three candidate responses were generated with the intention of representing agreement, disagreement, and balanced positions. In practice, model outputs did not always follow this scheme, especially on sensitive prompts where responses sometimes converged toward safe or hedged positions. We therefore treat responses $A$, $B$, and $C$ as empirical candidate responses rather than theoretically controlled stance conditions. The compression analysis concerns whether accepted non-winner responses are lost under single-winner aggregation, not whether a specific intended stance category wins.

\paragraph{Generalisability.}
Malaysia was chosen as a diagnostic setting because its structural plurality makes the measurement problem visible. Whether similar patterns arise in other plural societies, other domains, or larger annotation designs remains an empirical question. The present study establishes a diagnostic phenomenon and a measurement protocol for studying it. It does not claim to identify all conditions under which Preference-Validity Compression will occur.
%\section*{Acknowledgments}
%Omitted for anonymous review.

\clearpage
\newpage
\bibliography{custom}

\clearpage
\newpage
\appendix

\section{Proof of Proposition}
\label{app:proof}

Let $V(x)=\{y_1,\ldots,y_k\}$ be the set of plural-valid responses for prompt $x$, where $k>1$. Scalar aggregation assigns each $y_j\in V(x)$ an acceptance frequency
\begin{equation}
A(x,y_j)=\frac{1}{n}\sum_{i=1}^{n}u_i(x,y_j).
\end{equation}
The single-winner objective selects
\begin{equation}
y^* \in \arg\max_{y\in Y_x} A(x,y).
\end{equation}
The output is one selected response from $Y_x$. Since $|V(x)|>1$, at least one response $y_j\in V(x)$ satisfies $y_j\neq y^*$. By definition, this response remains plural-valid, but it is not preserved in the single-winner target. Therefore, the mapping from $V(x)$ to $y^*$ compresses a set of valid responses into one selected response.

\section{Participant Recruitment}
\label{app:recruitment}
\subsection{Recruitment and Selection}
In total, 53 individuals registered interest through an online form. From this pool, 44 were selected to provide coverage across Malay, Chinese, Indian, as well as indigenous communities from Sabah and Sarawak, across age bands from 18-24 to 65 and above. After task completion, quality 
filtering, and linkage to usable evaluation records, the final analytical 
sample contained 20 unique participants. Participants received financial 
compensation aligned with Malaysian minimum wage requirements.

\subsection{Sample Demographics}
The 20 participants include those from Malay, Chinese, Indian, and several 
East Malaysian indigenous backgrounds, including Bajau, Kadazan-Dusun, 
Kayan, and Melanau. Some demographic cells, especially older age bands, 
contain only one participant. Demographic variables are therefore used only 
as interpretive context, not as predictors of preference. The sample is not 
intended to estimate population-level Malaysian preference distributions. 
Table~\ref{tab:demographics} reports the full demographic breakdown.

The demographic questionnaire was available in Bahasa Melayu and English. 
Where applicable, Bahasa Melayu responses were translated into English for 
consistency in reporting.

\subsection{Research Ethics and Data Privacy}
All participants provided informed consent before beginning the task. All findings are reported in anonymised form.

Before beginning the task, participants reviewed and agreed to a consent statement that disclosed:
\begin{itemize}
    \item The purpose of the study and that responses would be used for research and any resulting publication.
    \item That personally identifiable information collected during registration would be kept separate from the analytical dataset and not reported.
    \item Compensation terms aligned with Malaysian minimum wage requirements.
    \item The right to withdraw at any time.
    \item Contact information for questions or concerns about the study.
\end{itemize}

\subsection{Content Advisory}
Participants were also informed in advance that some prompts touch on 
sensitive topics relevant to the Malaysian context, including race 
relations, religious practice, language policy, and regional identity. Participants were told they could skip any prompt they were uncomfortable evaluating, and could withdraw from the study at any time without forfeiting compensation for completed work.

\section{Participant Instructions and Task Framing}
\label{app:instructions}

This section reports the full instructions and task framing shown to participants during evaluation, in the order they appeared.

\subsection{Study Overview Shown to Participants}
Participants were introduced to the task with the following description:
\begin{quote}
You will read a user question and three different AI responses to that question. Your task is to decide whether each response is acceptable to you. 

There is no single correct answer. Different people may disagree on what is acceptable, and that is expected in this study. We are interested in your personal judgment and lived experience.
\end{quote}

\subsection{Rating Instructions}
For each prompt, participants were shown the prompt followed by three candidate responses (labelled A, B, and C). For each response, participants were asked to:
\begin{enumerate}
    \item Mark the response as either \textbf{acceptable} or \textbf{not acceptable} as a reply to the prompt in the Malaysian context.
    \item Provide a brief written justification for the rating.
\end{enumerate}
Participants were explicitly told that they could mark any number of responses as acceptable, including zero, one, two, or all three. They were instructed that there was no expected answer and no preference for a particular response position.

\section{Prompt Generation}
\label{sec:prompt-generation}

Prompt generation followed a three-stage pipeline designed to ground the  corpus in authentic Malaysian discourse rather than researcher assumptions  about local disagreement. Each stage is described in the subsections below.

\subsection{Data Collection and Processing}
\label{sec:data-collection}

\begin{table}[h]
\small
\setlength{\tabcolsep}{7pt}
\renewcommand{\arraystretch}{1.2}
\centering
\caption{Data Collection: Files and utterances per platform after segmentation, 
filtration, and deduplication}
\label{tab:data_stats}
\begin{tabular}{|l|r|r|}
\hline
\textbf{Platform} & \textbf{Files} & \textbf{Utterances (processed)} \\
\hline
Lowyat.net & 3,856 & 458,433 \\
Reddit & 5,665 & 153,097 \\
Threads & 6,739 & 74,419 \\
YouTube & 10,203 & 6,970 \\
\hline
\textbf{Total} & \textbf{26,463} & \textbf{692,919} \\
\hline
\end{tabular}
\end{table}

Forum discourse was collected from four major Malaysian online platforms, \textit{i.e.,} Lowyat.net, Reddit (Malaysian communities), Threads, and YouTube podcast channels. Raw data underwent multi-stage processing, such as segmentation into utterance-level chunks, filtration of off-topic content, and deduplication to remove redundant discourse. The processing pipeline produced 692,919 distinct utterances from 26,463 source files across all platforms (refer to Table~\ref{tab:data_stats}).

YouTube data was sourced from seven prominent Malaysian podcast channels, which are Morning Brief (BFM), Keluar Sekejap, Are We OK?, The Sambal Pod, Kinabalu Podcast, BuatSaja Podcast, and Yang Berhenti Menteri (hosted by Mohd Rafizi bin Ramli).

\subsection{Topic Identification}
\label{sec:topic-identification}

Topics were identified from the processed discourse using a systematic approach grounded in locally relevant cultural, linguistic, and policy dimensions. The resulting topic catalog spans 150 topics organized across three domains: Culture (55 topics), Language (64 topics), as well as Local governance and policy (31 topics). Each domain contains multiple thematic dimensions (refer to Table~\ref{tab:topics_catalog}).

\clearpage

\onecolumn
\renewcommand{\arraystretch}{1.15}
\begin{longtable}[c]{lrr}
\caption{Full demographic breakdown of the final analytical sample ($N=20$). Free-text entries (e.g., ``Kayan'' under \textit{Ethnicity} and ``Part Time Admin'' under \textit{Employment / Sector}) are reported as submitted.}
\label{tab:demographics} \\
\toprule
\textbf{Category} & \textbf{$n$} & \textbf{\%} \\
\midrule
\endfirsthead
\multicolumn{3}{l}{\textit{Total Participants}} \\
\quad Total & 20 & 100.0\% \\
\\[-6pt]

\multicolumn{3}{l}{\textit{Age}} \\
\quad 18--24 & 6 & 30.0\% \\
\quad 25--34 & 4 & 20.0\% \\
\quad 35--44 & 4 & 20.0\% \\
\quad 45--54 & 4 & 20.0\% \\
\quad 55--64 & 1 & 5.0\%  \\
\quad 65+    & 1 & 5.0\%  \\
\\[-6pt]

\multicolumn{3}{l}{\textit{Gender}} \\
\quad Female & 15 & 75.0\% \\
\quad Male   & 5  & 25.0\% \\
\\[-6pt]

\multicolumn{3}{l}{\textit{Ethnicity}} \\
\quad Chinese       & 9 & 45.0\% \\
\quad Malay         & 5 & 25.0\% \\
\quad Indian        & 2 & 10.0\% \\
\quad Bajau         & 1 & 5.0\%  \\
\quad Kadazan-Dusun & 1 & 5.0\%  \\
\quad Kayan$^{*}$   & 1 & 5.0\%  \\
\quad Melanau       & 1 & 5.0\%  \\
\\[-6pt]

\multicolumn{3}{l}{\textit{Religion}} \\
\quad Islam                          & 8 & 40.0\% \\
\quad Buddhism                       & 5 & 25.0\% \\
\quad Christianity                   & 3 & 15.0\% \\
\quad Hinduism                       & 2 & 10.0\% \\
\quad Taoism / Chinese folk religion & 1 & 5.0\%  \\
\quad No religion                    & 1 & 5.0\%  \\
\\[-6pt]

\multicolumn{3}{l}{\textit{Education}} \\
\quad Bachelor's degree              & 13 & 65.0\% \\
\quad STPM / A-level / Diploma       & 4  & 20.0\% \\
\quad Postgraduate (Master's or PhD) & 2  & 10.0\% \\
\quad SPM / O-level                  & 1  & 5.0\%  \\
\\[-6pt]

\multicolumn{3}{l}{\textit{Secondary School Type}} \\
\quad SMK (National secondary school)              & 14 & 70.0\% \\
\quad Chinese independent school                   & 2  & 10.0\% \\
\quad SMJK (C) (Chinese-medium national-type)      & 2  & 10.0\% \\
\quad Government boarding school (MRSM, SBP, etc.) & 1  & 5.0\%  \\
\quad International / private                      & 1  & 5.0\%  \\
\\[-6pt]

\multicolumn{3}{l}{\textit{Primary Language (Daily Use)}} \\
\quad English       & 7 & 35.0\% \\
\quad Mandarin      & 7 & 35.0\% \\
\quad Bahasa Melayu & 5 & 25.0\% \\
\quad Cantonese     & 1 & 5.0\%  \\
\\[-6pt]

\multicolumn{3}{l}{\textit{Employment / Sector}} \\
\quad Student                            & 5 & 25.0\% \\
\quad Engineering or technical services  & 3 & 15.0\% \\
\quad Government / public service        & 3 & 15.0\% \\
\quad Not currently working              & 3 & 15.0\% \\
\quad AI / data / tech sector            & 1 & 5.0\%  \\
\quad Business or management             & 1 & 5.0\%  \\
\quad Creative, media, and entertainment & 1 & 5.0\%  \\
\quad Part Time Admin$^{\dagger}$        & 1 & 5.0\%  \\
\quad Self-employed / own business       & 1 & 5.0\%  \\
\quad Wildlife conservation              & 1 & 5.0\%  \\
\\[-6pt]

\multicolumn{3}{l}{\textit{State Grown Up In}} \\
\quad Kuala Lumpur    & 3 & 15.0\% \\
\quad Sabah           & 3 & 15.0\% \\
\quad Penang          & 2 & 10.0\% \\
\quad Perak           & 2 & 10.0\% \\
\quad Sarawak         & 2 & 10.0\% \\
\quad Selangor        & 2 & 10.0\% \\
\quad Johor           & 1 & 5.0\%  \\
\quad Kedah           & 1 & 5.0\%  \\
\quad Kelantan        & 1 & 5.0\%  \\
\quad Negeri Sembilan & 1 & 5.0\%  \\
\quad Pahang          & 1 & 5.0\%  \\
\quad Terengganu      & 1 & 5.0\%  \\
\\[-6pt]

\multicolumn{3}{l}{\textit{State Currently Living In}} \\
\quad Selangor        & 6 & 30.0\% \\
\quad Kuala Lumpur    & 4 & 20.0\% \\
\quad Negeri Sembilan & 3 & 15.0\% \\
\quad Sarawak         & 2 & 10.0\% \\
\quad Johor           & 1 & 5.0\%  \\
\quad Kedah           & 1 & 5.0\%  \\
\quad Melaka          & 1 & 5.0\%  \\
\quad Putrajaya       & 1 & 5.0\%  \\
\quad Terengganu      & 1 & 5.0\%  \\

\end{longtable}
{\footnotesize $^{*}$Free-text entry submitted under the ``Other'' \textit{Ethnicity} option.}

{\footnotesize $^{\dagger}$Free-text entry submitted under the ``Other'' \textit{Employment / Sector} option.}
\twocolumn[
  \begin{@twocolumnfalse}
  \vspace{6pt}
\noindent
\begin{minipage}{\textwidth}
\small
\setlength{\tabcolsep}{8pt}
\renewcommand{\arraystretch}{1.3}
\centering
\captionof{table}{150 topics organized by domain (Culture, Language, 
Local) and thematic dimension}
\label{tab:topics_catalog}
\begin{tabular}{|p{2.8cm}|p{6.4cm}|r|}
\hline
\textbf{Topic ID} & \textbf{Dimension} & \textbf{Count of Topics} \\
\hline
\multicolumn{3}{|l|}{\textbf{Culture}} \\
T0004--T0054 & \textit{Values \& Norms:} Sensitivity \& Taboos & 10 \\
T0002--T0055 & \textit{Values \& Norms:} Ethical \& Moral Standards & 7 \\
T0041--T0053 & \textit{Context:} Historical \& Societal & 3 \\
T0033--T0051 & \textit{Belief Systems:} Religious Perspectives & 11 \\
T0001--T0052 & \textit{Values \& Norms:} Social Acceptability & 5 \\
T0022--T0048 & \textit{Symbolism:} Cultural Meaning & 2 \\
T0003--T0047 & \textit{Context:} Temporal \& Civic & 17 \\
\hline
\multicolumn{3}{|l|}{\textbf{Language}} \\
T0102--T0106 & \textit{Expression:} Formality \& Register & 5 \\
T0056--T0063 & \textit{Coverage:} Language Scope & 8 \\
T0111--T0114 & \textit{Interaction:} Multilingual & 4 \\
T0115--T0119 & \textit{Interpretation:} Pragmatic Understanding & 5 \\
T0107--T0110 & \textit{Variation:} Regional Dialects & 4 \\
T0064--T0101 & \textit{Lexicon:} Terminology, Slang, Local Expressions & 38 \\
\hline
\multicolumn{3}{|l|}{\textbf{Local}} \\
T0120--T0146 & \textit{Regulation:} Constitutional \& Legal Requirements & 11 \\
T0129--T0150 & \textit{Policy:} National Priorities \& Governance & 16 \\
T0132--T0144 & \textit{Regulation:} Sectoral Requirements & 4 \\
\hline
\end{tabular}
\vspace{6pt}
\end{minipage}
\vspace{12pt}
\end{@twocolumnfalse}
]
\subsection{Persona Development}
\label{sec:persona-development}

From the extracted topics, we developed 20 diverse persona profiles in three stages. (i) voice cluster extraction: identifying 1 to 3 distinct discourse voices per topic segment, each characterized by emotional posture, friction triggers, and phrasing patterns; (ii) profile bundling: clustering voice patterns by similarity using weighted similarity metrics (lexical overlap 55\%, emotional proximity 20\%, posture alignment 15\%, topic dimension match 10\%), with high-similarity clusters merged above a 0.55 threshold; and (iii) refinement: using an LLM to synthesize each bundle into a coherent persona with handle, core concern, and emotional profile (see Table~\ref{tab:persona_profiles}).

Each persona targets specific topics through its distinct emotional stance and friction triggers, enabling generation of locally grounded prompts across multiple valid perspectives.

\subsection{Quality Filtering}
\label{sec:quality-filtering}
\begin{figure}[h]
  \centering
  \includegraphics[width=\linewidth]{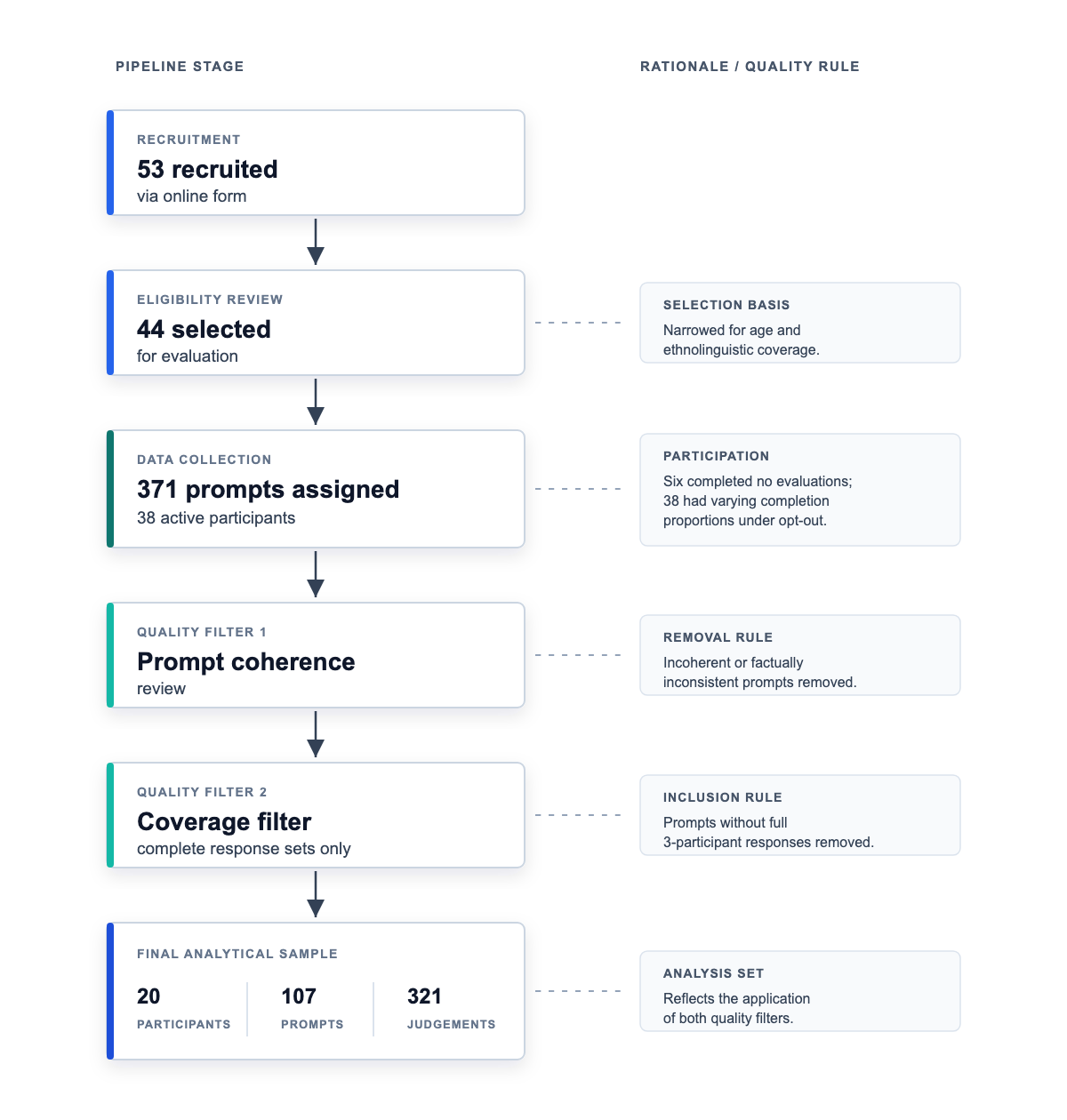}
  \caption{Data pipeline from recruitment to the final analytical sample.}
  \label{fig:study-design-flowchart}
  %\vspace{-1em}
\end{figure}

Two quality filters were applied after data collection. First, four members of the research team independently reviewed prompt coherence, with difficult cases discussed collectively. Prompts were removed when the scenario was internally inconsistent or factually incoherent in ways that made acceptability judgments uninterpretable rather than genuinely contestable. Second, a coverage filter removed prompts that did not receive responses from all three assigned participants. This ensured that every retained prompt supported a minimal two-out-of-three majority judgment. Because both filters were applied as an integrated process, filter-specific attrition is reported only at the aggregate level in Figure~\ref{fig:study-design-flowchart}.

\onecolumn
\begin{table*}[t]
\small
\setlength{\tabcolsep}{7pt}
\renewcommand{\arraystretch}{1.25}
\centering
\caption{Persona Profiles: 20 diverse personas grouped by primary concern, 
each with associated emotional posture and friction triggers}
\label{tab:persona_profiles}
\begin{tabular}{|p{0.85cm}|p{2.2cm}|p{3.0cm}|p{1.8cm}|p{3.5cm}|}
\hline
\textbf{ID} & \textbf{Handle} & \textbf{Core Concern} & \textbf{Emotion} 
& \textbf{Key Triggers} \\
\hline
\multicolumn{5}{|l|}{\textit{Accountability \& Justice}} \\
P001 & Outraged Citizen & Public safety \& accountability & outraged & 
Drunk driving, corruption, lack of action \\
P004 & Skeptical Watchdog & Institutional accountability & skeptical & 
Official statements, safety procedures \\
P008 & Scammed Citizen & Fraud \& justice & frustrated & Scamming, 
ineffective law enforcement \\
P017 & Accountable Parent & Parental responsibility & assertive & Blaming 
others, negligence \\
\hline
\multicolumn{5}{|l|}{\textit{Patriotism \& National Identity}} \\
P003 & Indignant Patriot & National pride & indignant & Disrespect for 
flag, disloyalty \\
P010 & Angry Nationalist & Economic exploitation & angry-resentful & 
Disrespect, national symbols, price hikes \\
P011 & Disillusioned Patriot & Ethnic identity & sarcastic & Cultural 
misidentification, exclusion \\
P019 & Passionate Patriot & National language & passionate & 
Underestimation, dismissal \\
P020 & Sovereign Advocate & State rights & defiant & Taking control, 
defending rights \\
\hline
\multicolumn{5}{|l|}{\textit{Morality \& Standards}} \\
P002 & Moral Guardian & Moral outrage & angry & Irresponsible behavior, 
disrespect \\
P006 & Prejudice Propagator & Stereotyping & aggressive & Minor incidents, 
stereotyping \\
P009 & Social Justice Advocate & Labor rights & indignant & Salary cuts, 
forced resignations \\
\hline
\multicolumn{5}{|l|}{\textit{Skepticism \& Cynicism}} \\
P007 & Cynical Watchdog & Political corruption & cynical & Political 
excuses, blame game \\
P012 & Truth Seeker & Hidden truths & suspicious & Suppression of truth, 
hidden agendas \\
P013 & Cynical Observer & Hypocrisy & cynical & Hypocrisy, double 
standards \\
P016 & Systemic Critic & Systemic bias & resentful & Racial division, 
political rhetoric \\
\hline
\multicolumn{5}{|l|}{\textit{Pragmatism \& Concern}} \\
P005 & Diplomatic Pragmatist & International relations & concerned & 
Conflicting alliances, diplomatic issues \\
P014 & Privacy Watchdog & Digital privacy & anxious & Identity control, 
surveillance \\
P015 & Curious Inquirer & Understanding & curious & Lack of context, 
unexplained claims \\
P018 & Frustrated Malaysian & Fair pay & frustrated & Low salary, 
dismissive employers \\
\hline
\end{tabular}
\end{table*}
\clearpage
\newpage

\twocolumn
\section{Response Generation}
\label{sec:response-generation}
To generate diverse responses across multiple valid perspectives, we implemented stance-controlled response generation by injecting three distinct reasoning frameworks into three separate language models. Each prompt and response was then assigned a stance tag by five members of the research team using a schema developed for this study. They are used to make interpretive frames visible and to support qualitative interpretation, not to support the main quantitative claims.

\subsection{Response Generation Models}

Three LLMs were assigned one stance each to generate diverse responses across all prompts:

\begin{itemize}
\item \textbf{Response A (Agree):} DeepSeek-V4-Flash
\item \textbf{Response B (Disagree):} GPT-5-Nano
\item \textbf{Response C (Diplomatic):} Gemini-3.1-Flash-Lite-Preview
\end{itemize}

This assignment ensures that for each controversial prompt, annotators encounter one response reinforcing the position, one challenging it, and one analyzing the underlying dilemma, which enabling assessment of acceptability across legitimate but distinct interpretive frames.    

One practical deviation from the intended design is worth noting. The model outputs did not consistently follow the assigned stance orientation, particularly on sensitive prompts where responses sometimes converged toward hedged or non-committal positions regardless of the injected reasoning framework. For this reason, the main analysis treats response positions $A$, $B$, and $C$ as empirical candidate responses rather than theoretically controlled stance conditions.

\subsection{Response Stances Definitions}

\label{sec:response-stances-definition}

The three stances were operationalized as follows:

\textbf{Agree:} Model adopts and reinforces a given position by building upon its logic with supporting evidence and shared principles. Rather than passive agreement, this stance actively strengthens the reasoning through deeper justifications and a confirmatory tone.

\textbf{Disagree:} Model presents an alternative perspective grounded in competing priorities and values. Rather than dismissal, this stance constructs a counter-argument that acknowledges complexity while maintaining a principled boundary against the original position.

\textbf{Diplomatic:} Model analyzes the underlying tension between competing values without advocating for either side. This stance decomposes the dilemma into its constituent trade-offs, presenting both gains and losses across scenarios in neutral, balanced language.

\section{Stance Labeling}
\label{sec:stance-schema}
Each prompt and response was assigned a stance tag from a schema developed for this study. Labeling was conducted independently by five members of the research team. No reconciliation procedure was applied, and disagreements between labelers were not adjudicated into a single resolved label. This choice reflects the descriptive role of the stance labels. They are used as characterizations of response content rather than ground-truth categorical measurements. The main quantitative results rely on acceptability counts, not on stance-label agreement.

Examples of stance labels include \textit{pro-multilingual-inclusive-education}, \textit{pro-Malay-language-culture}, and \textit{anti-LGBTQ-symbols}. The labels characterize value orientations embedded in prompts and responses. They are not used to classify participants ideologically. Refer to Table~\ref{tab:prompt_stances} for the full list of unique prompt stance labels and Table~\ref{tab:response_stances} for the full list of unique response stance labels.
\clearpage
\newpage

\onecolumn
\begin{longtable}{|p{5cm}|r|}
\caption{All 116 Unique Prompt Stances}
\label{tab:prompt_stances}\\
\hline
\textbf{Stance} & \textbf{Count} \\
\hline
\endfirsthead
\multicolumn{2}{c}{\textit{Continued from previous page}} \\
\hline
\textbf{Stance} & \textbf{Count} \\
\hline
\endhead
\hline
\multicolumn{2}{r}{\textit{Continued on next page}} \\
\endfoot
\hline
\endlastfoot
\footnotesize
\setlength{\tabcolsep}{8pt}
\renewcommand{\arraystretch}{1.1}
pro-regional-dialect & 32 \\
anti-bureaucratic-neglect & 28 \\
protect-privacy & 20 \\
anti-institution & 19 \\
anti-selective-enforcement & 9 \\
parental-accountability-gap & 9 \\
anti-surveillance & 8 \\
anti-impunity-for-the-powerful & 6 \\
anti-workplace-discrimination & 6 \\
cross-cultural-inquiry & 6 \\
demand-accountability-ordinary-malaysians & 6 \\
pro-accountability-home-parenting & 6 \\
pro-national-language & 6 \\
pro-parental\_accountability & 5 \\
protect-workers & 5 \\
east-malaysian-autonomy-inquiry & 3 \\
advocate-parental-accountability & 3 \\
anti-chinese & 3 \\
anti-indian & 3 \\
anti-blaming-parents & 3 \\
anti-bureaucracy-political-rhetoric & 3 \\
anti-cooptation-culture & 3 \\
anti-corporate-restriction & 3 \\
anti-cover-up-staff-misconduct & 3 \\
anti-deception-dialect\_identity & 3 \\
anti-disrespect-islam & 3 \\
anti-ekyc-data-collection & 3 \\
anti-exploitation & 3 \\
anti-exploitation-workplace & 3 \\
anti-government-negligence & 3 \\
anti-intervention---federal-government & 3 \\
anti-leniency-gisb & 3 \\
anti-manipulasi-bahasa-sopan & 3 \\
anti-manipulation-of-religion & 3 \\
anti-politicisation-of-religious-enforcement & 3 \\
anti-politicisation-religion & 3 \\
anti-politicization-transliteration & 3 \\
anti-surveillance-mykad & 3 \\
anti-worker-exploitation & 3 \\
anti-workplace-exploitation & 3 \\
anti-workplace\_discrimination & 3 \\
balance-accountability-with-religious-principles & 3 \\
cultural-legitimacy-consistency-critique & 3 \\
defend-islam & 3 \\
defend-halal-integrity & 3 \\
demand-accountability-governance & 3 \\
demand-clarity-in-bureaucratic-communication & 3 \\
demand-justice-with-integrity & 3 \\
demand-systemic-accountability-youth\_outcomes & 3 \\
demand-transparency-in-safeguarding-oversight & 3 \\
demand-transparency-platform\_moderation & 3 \\
governance-compliance-inquiry & 3 \\
heritage\_narratives-inquiry & 3 \\
inquire-parental\_responsibility & 3 \\
inquiry-language\_identity & 3 \\
language-norms-double-standard-inquiry & 3 \\
liability\_clarity\_enquiry & 3 \\
multilingual-codeswitching-inquiry & 3 \\
preserve-national\_language & 3 \\
pro-accountability-in-religious-leadership & 3 \\
pro-citizenship-law-reform-for-children-born-to-malaysian-mothers-abroad & 3 \\
pro-due\_diligence & 3 \\
pro-parental-accountability & 3 \\
pro-personal-accountability-and-normative-internalisation-of-responsibility & 3 \\
pro-privacy-protection & 3 \\
pro-regional-autonomy & 3 \\
pro-religious-inclusivity & 3 \\
pro-respect-religious-terms & 3 \\
pro-space-design-accountability & 3 \\
pro-standardization---address-abbreviations & 3 \\
pro-vernacular\_preservation & 3 \\
promote-consistent-enforcement & 3 \\
protect-islam & 3 \\
protect-regional-dialect & 3 \\
religious-economic-governance-trade-off-evaluation & 3 \\
anti-federal-centralization & 2 \\
pro-islamic\_term\_protection & 2 \\
advocate-remote-work & 2 \\
anti-lgbtq-symbols & 2 \\
anti-authority & 2 \\
anti-blame-shifting-accountability & 2 \\
anti-corruption & 2 \\
anti-empty-rhetoric & 2 \\
anti-evading-responsibility & 2 \\
anti-gimmick-labeling-musang-king & 2 \\
anti-labour-exploitation & 2 \\
anti-manipulation-code-switching & 2 \\
anti-non-malays & 2 \\
anti-religious-hypocrisy & 2 \\
anti-rulebreaking-traffic-regulations & 2 \\
anti-silence-from-religious-bodies & 2 \\
anti-surveillance-ekyc & 2 \\
defend-malay-dignity & 2 \\
degree-career-mismatch-inquiry & 2 \\
demand-transparency-in-official-communication & 2 \\
inquire-parental\_responsibility\_in\_cancel\_culture & 2 \\
neutral-ma63-inquiry & 2 \\
perceived-enforcement-inequality & 2 \\
pro-chinese-malaysian-heritage & 2 \\
pro-malay-language & 2 \\
pro-malay-language-culture & 2 \\
pro-diversity\_inclusion & 2 \\
pro-employee-voice & 2 \\
pro-heritage-language & 2 \\
pro-host-accountability & 2 \\
pro-malay-language & 2 \\
pro-national\_language & 2 \\
pro-parental-empowerment & 2 \\
pro-parental-accountability & 2 \\
pro-regional-identity & 2 \\
pro-standardisation-islamic-certification & 2 \\
pro-traditional-values & 2 \\
pro-transparency-in-governance & 2 \\
reject-derogatory-metaphor & 2 \\
religious-inclusivity-inquiry & 2 \\
uphold-religious-principles-in-diplomacy & 2 \\
\end{longtable}
\clearpage

\onecolumn
\begin{longtable}{|p{4.5cm}|r|c|}
\caption{All 127 Unique Response Stances} \label{tab:response_stances}\\
\hline
\textbf{Stance} & \textbf{Occ.} & \textbf{Topics} \\
\hline
\endfirsthead
\multicolumn{3}{c}{\textit{Continued from previous page}} \\
\hline
\textbf{Stance} & \textbf{Occ.} & \textbf{Topics} \\
\hline
\endhead
\hline
\multicolumn{3}{r}{\textit{Continued on next page}} \\
\endfoot
\hline
\endlastfoot
\small
\setlength{\tabcolsep}{8pt}
\renewcommand{\arraystretch}{1.1}
pluralist & 404 & 23 \\
pragmatist & 310 & 22 \\
institutionalist & 176 & 15 \\
activist & 15 & 4 \\
neutral-pluralist & 14 & 2 \\
anti-institution & 13 & 2 \\
pro-regional-dialect & 13 & 2 \\
diplomatic & 8 & 2 \\
anti-surveillance & 6 & 2 \\
pro-institutional-harmony- / -societal-unity & 6 & 2 \\
pro-local-resilience-pragmatist & 6 & 2 \\
pro-national-language & 6 & 1 \\
anti-selective-enforcement & 5 & 1 \\
pro-parental\_accountability & 5 & 2 \\
anti-impunity-for-the-powerful & 3 & 1 \\
anti-digital-parenting & 3 & 1 \\
anti-discriminatory-language & 3 & 1 \\
anti-institution & 3 & 1 \\
assimilationist & 3 & 1 \\
authenticity-vs-institutional-visibility-trade-off & 3 & 1 \\
balance-standardization-with-cultural-identity-pragmatist & 3 & 1 \\
communal-institutionalist & 3 & 1 \\
corporate-roi-centric-pragmatist & 3 & 1 \\
economic-pragmatist & 3 & 1 \\
efficiency-vs-child-development-pluralist & 3 & 1 \\
governance-design-pragmatist & 3 & 1 \\
liberal-pragmatist & 3 & 1 \\
national-language-primacy-institutionalist & 3 & 1 \\
neutral-pluralist & 3 & 1 \\
prioritize-national-language-policy-institutionalist & 3 & 1 \\
pro-community-governance-pragmatist & 3 & 1 \\
pro-hindu-traditions & 3 & 1 \\
pro-pantai-timur-regional-identity & 3 & 1 \\
pro-sabah-language-education-inclusion-pragmatist & 3 & 1 \\
pro-sabah-language-inclusion-pragmatist & 3 & 1 \\
pro-sabah-linguistic-identity & 3 & 1 \\
pro-accountability-home-parenting & 3 & 1 \\
pro-accountability-in-religious-leadership & 3 & 1 \\
pro-bureaucratic-accountability-and-efficiency-pragmatist & 3 & 1 \\
pro-community-empowerment & 3 & 1 \\
pro-community-governance-pragmatist & 3 & 1 \\
pro-community-involvement-pragmatist & 3 & 1 \\
pro-consistent-enforcement-pragmatist & 3 & 1 \\
pro-governance-accountability-pragmatist & 3 & 1 \\
pro-heritage\_recognition & 3 & 1 \\
pro-hybrid-pathways-pragmatist & 3 & 1 \\
pro-individual-and-parental-accountability & 3 & 1 \\
pro-individual-cultural-ownership & 3 & 1 \\
pro-individual-responsibility & 3 & 1 \\
pro-institutional-cultural-stewardship & 3 & 1 \\
pro-institutional-enforcement & 3 & 1 \\
pro-integration-pragmatist & 3 & 1 \\
pro-labur-reform-pragmatist & 3 & 1 \\
pro-moral-stability & 3 & 1 \\
pro-multicultural-understanding & 3 & 1 \\
pro-multilingual-inclusive-education & 3 & 1 \\
pro-parental-accountability & 3 & 1 \\
pro-parental-accountability-pragmatist & 3 & 1 \\
pro-privacy-pragmatist & 3 & 1 \\
pro-religious-governance-integration & 3 & 1 \\
pro-social-cohesion & 3 & 1 \\
pro-social-infrastructure-for-parental-accountability & 3 & 1 \\
pro-space-redesign & 3 & 1 \\
pro-state-enforcement-institutionalist & 3 & 1 \\
pro-state-institutional-legitimacy & 3 & 1 \\
pro-state-mediated-religious-moral-governance & 3 & 1 \\
pro-structural-accountability-pragmatist & 3 & 1 \\
pro-system-authority-institutionalist & 3 & 1 \\
pro-system-efficiency-pragmatist & 3 & 1 \\
pro-traditional-family-values & 3 & 1 \\
pro-transparency-in-safeguarding-oversight & 3 & 1 \\
pro-workplace-fairness & 3 & 1 \\
promote-technocratic-cultural-governance-framework-pragmatist & 3 & 1 \\
situational-bilingualism-pluralist & 3 & 1 \\
structured-family-accountability-training-pragmatist & 3 & 1 \\
systemic-causality-framing & 3 & 1 \\
technocratic-integrity-governance-pragmatist & 3 & 1 \\
anti-top-down-cultural-legitimacy & 3 & 1 \\
anti-blaming-parents & 3 & 1 \\
anti-bureaucratic-neglect & 3 & 1 \\
anti-government-negligence & 3 & 1 \\
anti-leniency-gisb & 3 & 1 \\
anti-manipulasi-bahasa-sopan & 3 & 1 \\
anti-performative-politics & 3 & 1 \\
anti-politicization-transliteration & 3 & 1 \\
anti-regional-dialect & 3 & 1 \\
moral-sovereignty-based-on-local-values & 3 & 1 \\
pragmatic-balance & 3 & 1 \\
pro-sabah-autonomy-for-economic-efficiency & 3 & 1 \\
pro-employer-authority & 3 & 1 \\
pro-ethical-governance-(new) & 3 & 1 \\
pro-governance-compliance-pragmatism & 3 & 1 \\
pro-managed-linguistic-pluralism-pragmatist & 3 & 1 \\
pro-market-compliance-\&-efficiency & 3 & 1 \\
pro-multilingualism & 3 & 1 \\
pro-parents-accountability & 3 & 1 \\
pro-personal-accountability & 3 & 1 \\
pro-regional-autonomy & 3 & 1 \\
pro-respect-islam & 3 & 1 \\
pro-traditional-education-legitimacy & 3 & 1 \\
epistemic-methodological-pragmatist & 2 & 1 \\
individualistic & 2 & 1 \\
legal-empirical-institutionalist & 2 & 1 \\
pro-islamic\_term\_protection & 2 & 1 \\
pro-malay-dignity & 2 & 1 \\
pro-sabahan-malay-inclusion & 2 & 1 \\
pro-communal-trust & 2 & 1 \\
pro-multicultural-malaysian-identity & 2 & 1 \\
pro-privacy-\&-system-skeptical & 2 & 1 \\
pro-regional-rights & 2 & 1 \\
pro-regulatory-risk-balancing-pragmatist & 2 & 1 \\
pro-religious-integrity-in-diplomacy & 2 & 1 \\
pro-structured-tradition & 2 & 1 \\
anti-lgbtq-symbols & 2 & 1 \\
anti-authority & 2 & 1 \\
anti-labour-exploitation & 2 & 1 \\
anti-religious-hypocrisy & 2 & 1 \\
anti-standardisation-islamic-certification & 2 & 1 \\
anti-surveillance-ekyc & 2 & 1 \\
pro-malay-language-culture & 2 & 1 \\
pro-muslim-friendly-inclusivity & 2 & 1 \\
pro-community-conscience & 2 & 1 \\
pro-enforcement-accountability-pragmatist & 2 & 1 \\
pro-malay-language & 2 & 1 \\
pro-traditional-values & 2 & 1 \\
pro-transparency-in-governance & 2 & 1 \\
sympathetic-to-parents & 2 & 1 \\
\end{longtable}
\clearpage
\twocolumn
\section{Participant-Mode Diagnostic}
\label{app:participant-mode}
Figure~\ref{fig:consensus-rate-vs-breadth} reports the participant-level check used to test whether multi-selection reflects indiscriminate acceptance. For each participant, we compute selection breadth as the average number of accepted responses per prompt, and consensus alignment rate as the share of that participant's acceptances landing on responses that reached $\geq 2/3$ trio consensus.

\noindent
\begin{minipage}{\linewidth}
  \centering
  \includegraphics[width=\linewidth]{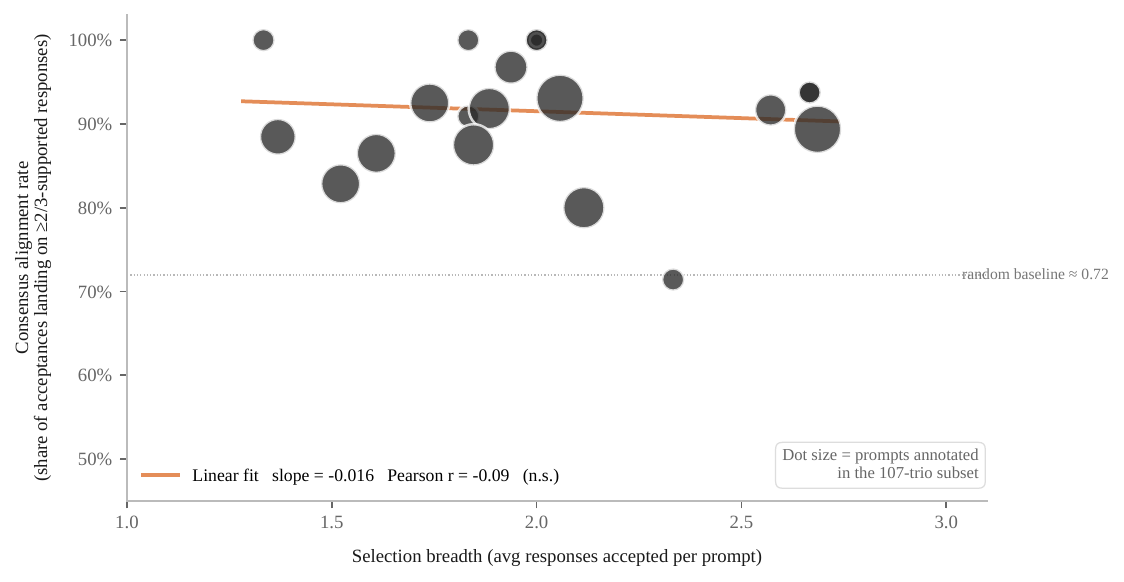}
  \captionof{figure}{Per-participant consensus alignment rate against 
  selection breadth for the 20 participants in the 107-trio subset. The 
  $y$-axis measures the share of each participant's acceptances that 
  landed on $\geq 2/3$ majority-supported responses. The slope is not 
  statistically significant, with $r=-0.09$ and $p=0.71$.}
  \label{fig:consensus-rate-vs-breadth}
\end{minipage}
\vspace{6pt}

Participant-level selection breadth ranges from 1.33 to 2.69 responses per prompt, while consensus alignment rates range from 71\% to 100\%. The absence of a detectable negative association suggests that broader selection is not simply lower-quality or indiscriminate acceptance. Instead, participants who accept more responses often surface additional majority-supported responses that single-winner aggregation leaves behind.
\section{Additional Result Visualizations}
\label{app:additional-results}

\noindent
\begin{minipage}{\linewidth}
  \centering
  \includegraphics[width=0.6\linewidth]{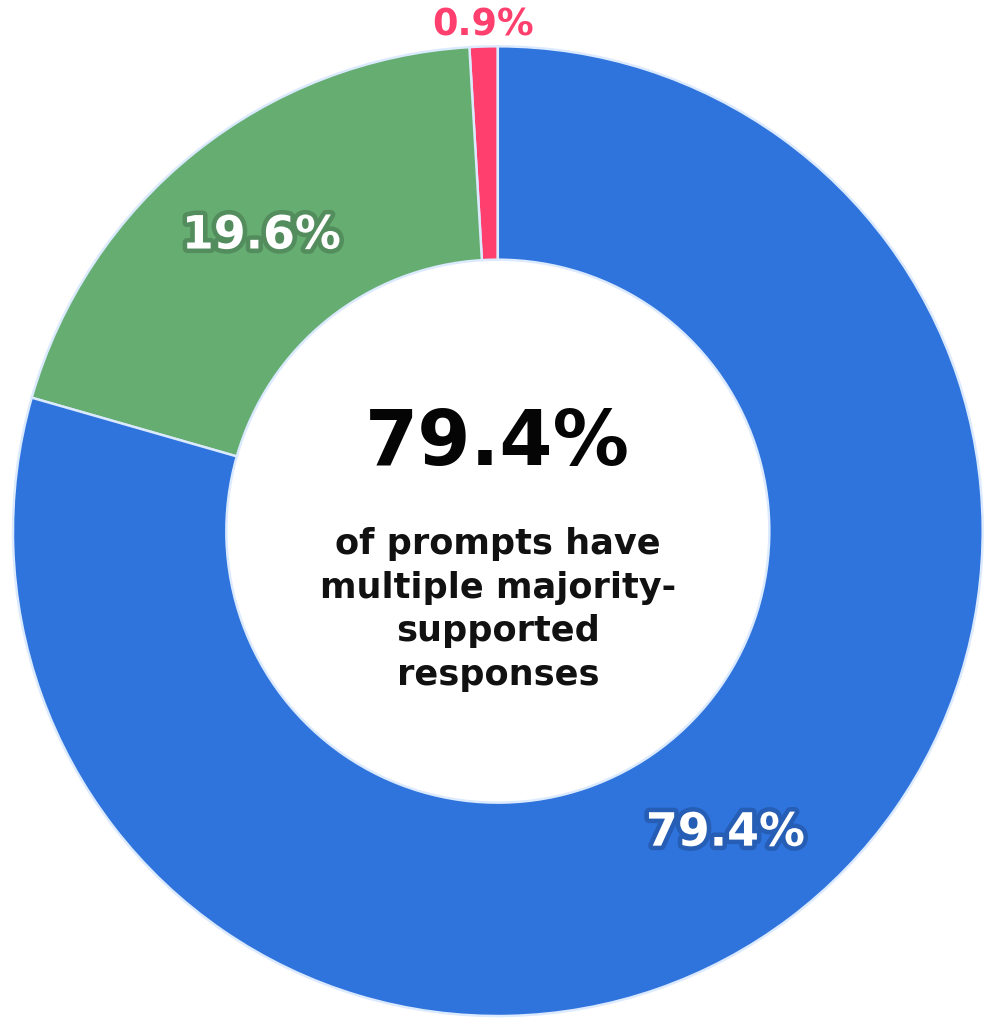}
  \captionof{figure}{Prompt-level prevalence of hidden consensus. Across 
  107 trio-annotated prompts, 79.4\% contain more than one response 
  reaching the $\geq 2/3$ majority acceptance threshold. Only 19.6\% 
  contain a single majority-supported response, and 0.9\% contain no 
  majority-supported response.}
  \label{fig:hidden-consensus-donut}
\end{minipage}
\vspace{6pt}

Figure~\ref{fig:hidden-consensus-donut} summarizes the prevalence of hidden consensus at the prompt level. The dominant pattern is not single-response convergence. Instead, most prompts contain multiple responses reaching the majority acceptance threshold, showing that the supported acceptability set is often non-singleton before aggregation.

\begin{figure}[h]
  \centering
  \includegraphics[width=\linewidth]{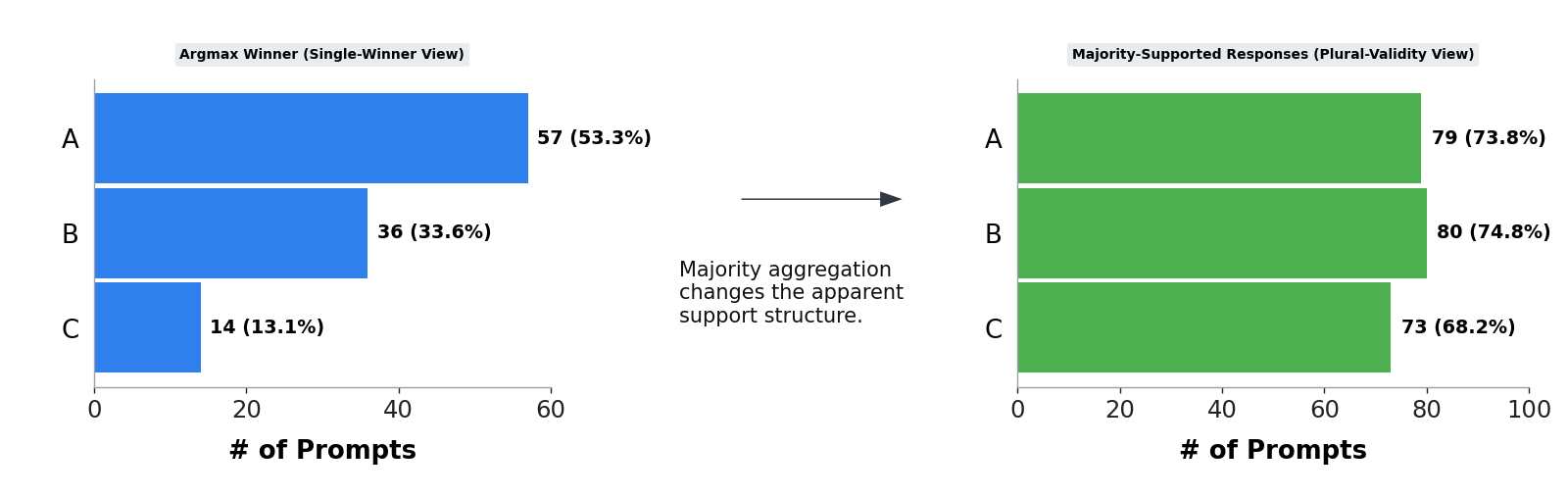}
  \caption{Ranking reversal under single-winner aggregation. The $\arg\max$ view ranks responses as $A>B>C$, with $A$ appearing as the dominant winner. When all majority-supported responses are counted, the ranking changes to $B>A>C$. This shows that single-winner aggregation changes the apparent support structure rather than merely summarizing it.}
  \label{fig:ranking-reversal}
  \vspace{-10pt}
\end{figure}

Figure~\ref{fig:ranking-reversal} provides an alternative visualization of the ranking reversal reported in Section~\ref{sec:results}. It is included to make clear that $\arg\max$ aggregation changes the apparent ordering of response support, not only the amount of support retained.

\end{document}